\documentclass[sigconf]{acmart}
\usepackage{bm}
\AtBeginDocument{%
  \providecommand\BibTeX{{%
    \normalfont B\kern-0.5em{\scshape i\kern-0.25em b}\kern-0.8em\TeX}}}

\setcopyright{acmcopyright}
\copyrightyear{2020}
\acmYear{2020}
\acmDOI{10.1145/nnnnnnn.nnnnnnn}

\acmPrice{15.00}
\acmISBN{978-x-xxxx-xxxx-x/xx/xx}

\begin{document}

\title{Contrastive Pre-training for Imbalanced Corporate Credit Ratings}

\orcid{1234-5678-9012}




\author{Bojing Feng}
\email{bojing.feng@cripac.ia.ac.cn}
\affiliation{%
  \institution{Center for Research on Intelligent Perception and Computing \and National Laboratory of Pattern Recognition \and Institute of Automation, Chinese Academy of Science}
}

\author{Wenfang Xue}
\authornote{Corresponding author}
\email{wenfang.xue@ia.ac.cn}
\affiliation{%
  \institution{Center for Research on Intelligent Perception and Computing \and National Laboratory of Pattern Recognition \and Institute of Automation, Chinese Academy of Science}
}



\begin{abstract}
  The corporate credit rating reflects the level of corporate credit and plays a crucial role in modern financial risk control. But real-world credit rating data usually shows long-tail distributions, which means a heavy class imbalanced problem challenging the corporate credit rating system greatly. To tackle that, inspired by the recent advances of pre-train techniques in self-supervised representation learning, we propose a novel framework named \textbf{C}ontrastive \textbf{P}re-training for \textbf{C}orporate \textbf{C}redit \textbf{R}ating (\textbf{CP4CCR}), which utilizes the self-supervision for getting over the class imbalance. Specifically, we propose to, in the first phase, exert contrastive self-supervised pre-training without label information, which aims to learn a better class-agnostic initialization. Furthermore, two self-supervised tasks are developed within \textbf{CP4CCR}: (i) \textbf{F}eature \textbf{M}asking (\textbf{FM}) and (ii) \textbf{F}eature \textbf{S}wapping(\textbf{FS}). In the second phase, we can train any standard corporate credit rating model initialized by the pre-trained network. Extensive experiments conducted on the real public-listed corporate rating dataset, prove that CP4CCR can improve the performance of standard corporate credit rating models, especially for the class with few samples.
\end{abstract}





\maketitle
\section{Introduction}\label{section:introduction}
Credit rating systems have been widely employed in all kinds of financial institutions such as banks and agency securities, to help them to mitigate credit risk. However, the credit assessment process, which is expensive and complicated, often takes months with experts involved to analyze. Therefore, it is necessary to design a model to predict credit level automatically.

The banking industry has developed some credit risk models since the middle of the twentieth century. Initially, logistic regression and hidden Markov model were applied in credit rating \cite{gogas2014forecasting,petropoulos2016novel}. With the progress of science and technology, machine learning, deep learning and hybrid models have shown their power in this field \cite{wu2012credit,yeh2012hybrid,pai2015credit}. With the advent of graph neural networks, some graph-based models \cite{bruss2019deeptrax,cheng2019dynamic,cheng2019risk,cheng2020contagious,cheng2020spatio}, were built based on the loan guarantee network to utilize relations between corporations.

Although these models have achieved promising results, they all ignore a fatal problem, the class imbalance problem, which prevents these models from improving continuously. We find out that credit rating agencies, like Standard \& Poor's, Moody's and CCXI, often give AA, A, BBB to most corporations, and few of them are CC, C. This will incur the class-imbalanced problem, which poses a great challenge to corporate credit rating model.

When the supervised model is limited, these models may fail due to the class-imbalanced problem, self-supervised manners will show their power. Recently, contrastive self-supervised learning brings a new chance to this problem. In the work \cite{yang2020rethinking}, they systematically investigate and demonstrate, theoretically and empirically, the class-imbalanced learning can significantly benefit in contrastive self-supervised manners. 

Contrastive self-supervised learning attracted the interest of many researchers due to its ability to avoid the cost of annotating large-scale datasets. Mikolov et al. \cite{mikolov2013distributed} firstly proposed contrastive learning for natural language processing in 2013. And it started to prevail on several NLP tasks in recent years \cite{chi2020infoxlm,fang2020cert,giorgi2020declutr}. On the one hand, contrastive self-supervised learning had shown its power by four main pretext tasks: color transformation, geometric transformation, context-based tasks, and cross-modal tasks \cite{jaiswal2020survey}. On the other hand, in the field of recommendation systems, Yao et al. \cite{Yao2020SelfsupervisedLF} proposed a multi-task self-supervised learning framework for large-scale recommendations. Xie et al. \cite{xie2020contrastive} developed CP4Rec to utilize the contrastive pre-training framework to extract meaningful sequential information. Besides, it can also alleviate generalization error, spurious correlations and adversarial attacks. That's why this kind of technology is applied in many fields.

Due to the limitations mentioned before, the application of contrastive self-supervised learning in the corporate credit rating is less well study. Different from previous works, we aim to tackle the class-imbalanced problem and improve the model performance of the class with few samples by leverage self-supervised signals constructed from corporate profile data. In specific, we propose a novel framework named Contrastive Pre-training for Corporate Credit Rating, CP4CCR for brevity. In the pre-training phase, we propose two self-supervised tasks, (i) Feature Masking and (ii) Feature Swapping (FS), to perform self-supervised pre-training. Then in the following phase, any standard corporate credit rating model can perform based on the pre-trained data. To sum up, the main contributions of this work are summarized as follows:

\begin{itemize}
    \item[$\bullet$]Two novel unsupervised tasks of corporate credit rating are proposed to perform contrastive pre-training: (i) Feature Masking and (ii) Feature Swapping (FS). 
    \item[$\bullet$]We proposed a new framework named \textbf{C}ontrastive \textbf{P}re-training for \textbf{C}orporate \textbf{C}redit \textbf{R}ating  (\textbf{CP4CCR}) to tackle the class-imbalanced problem.
    \item[$\bullet$]Comprehensive experiments on the real public-listed corporate rating dataset demonstrate that our model can improve the performance of common credit rating models experimentally.
\end{itemize}

\section{The Proposed Method: CP4CCR}\label{section:TheProposedMethod}
\begin{figure}[htbp]  
\centering  
\includegraphics[scale=0.21]{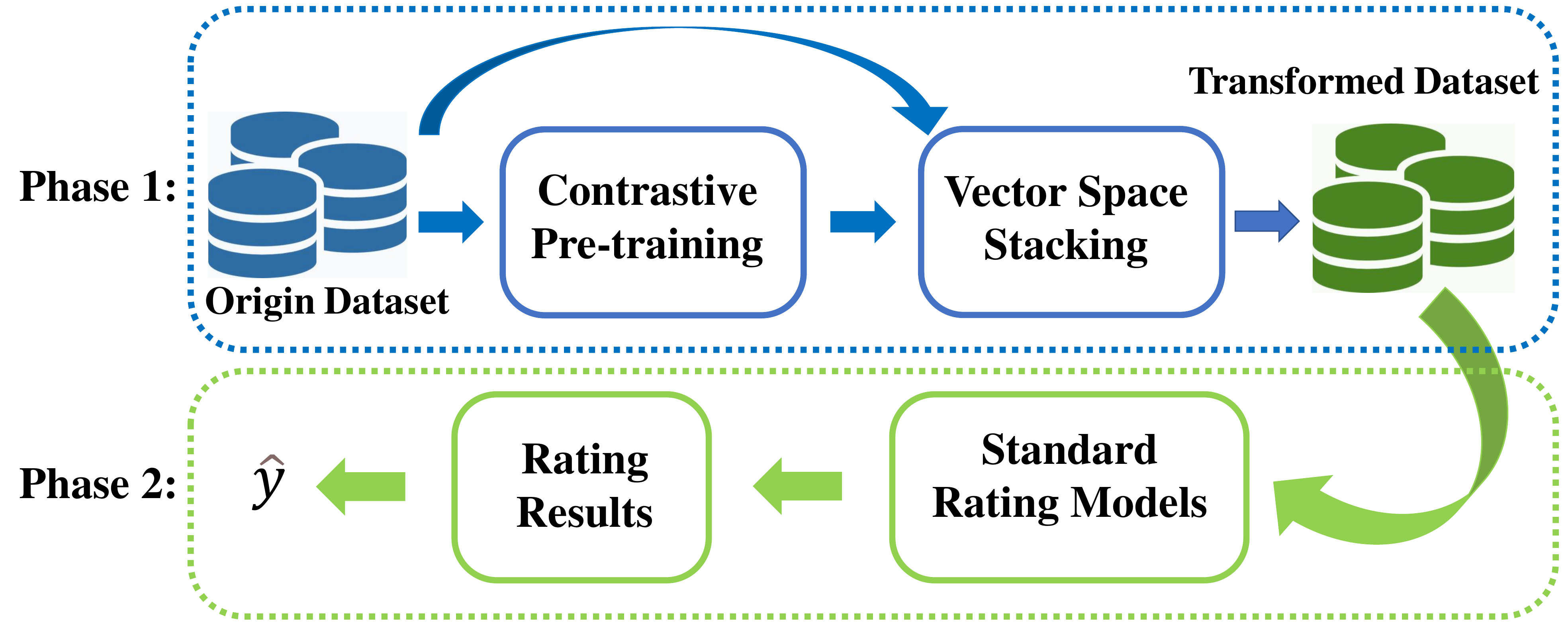}  
\caption{The architecture of the CP4CCR framework.} 
\label{figure1}
\end{figure}

In this section, we introduce the proposed framework: CP4CCR. It includes two phases. The architecture is illustrated in Figure \ref{figure1}.

In the first phase, the original dataset without label information is performed contrastive self-supervised pre-training to learn better initialization. Then to make full use of the original and pre-trained dataset, vector space stacking is proposed. We get the better-transformed dataset at the end of phase 1. Afterward, any standard corporate credit rating model can be applied based on the transformed dataset to predict the credit level in phase 2. 
\subsection{Notations and Problem Statement}
In this paper, we represent column vectors and matrices by italic lower case letters (e.g., $c$) and bold upper case letters (e.g., \bm{$X$}), respectively. And we use calligraphic letters to represent sets (e.g., $\mathcal{C}$).

Let $\mathcal{C}=\{c_1,c_2,\cdots,c_n\}$ denotes the set of corporations, $n$ is the number of corporations. $c\in \mathbb{R}^d$ includes the profile of corporation, which has $d$ dimensions feature. Every corporation has a corresponding label that represents its credit level. Let $\mathcal{Y}=\{y_1,y_2,\cdots,y_m\}$ denotes the set of labels, and  $m$ is the number of all unique labels. Corporate credit rating models aim to predict the credit level $\hat{y}$ according the profile of corporation $c$.
\subsection{Contrastive Self-supervised Pre-training}
\begin{figure}[htbp]  
\centering  
\includegraphics[scale=0.35]{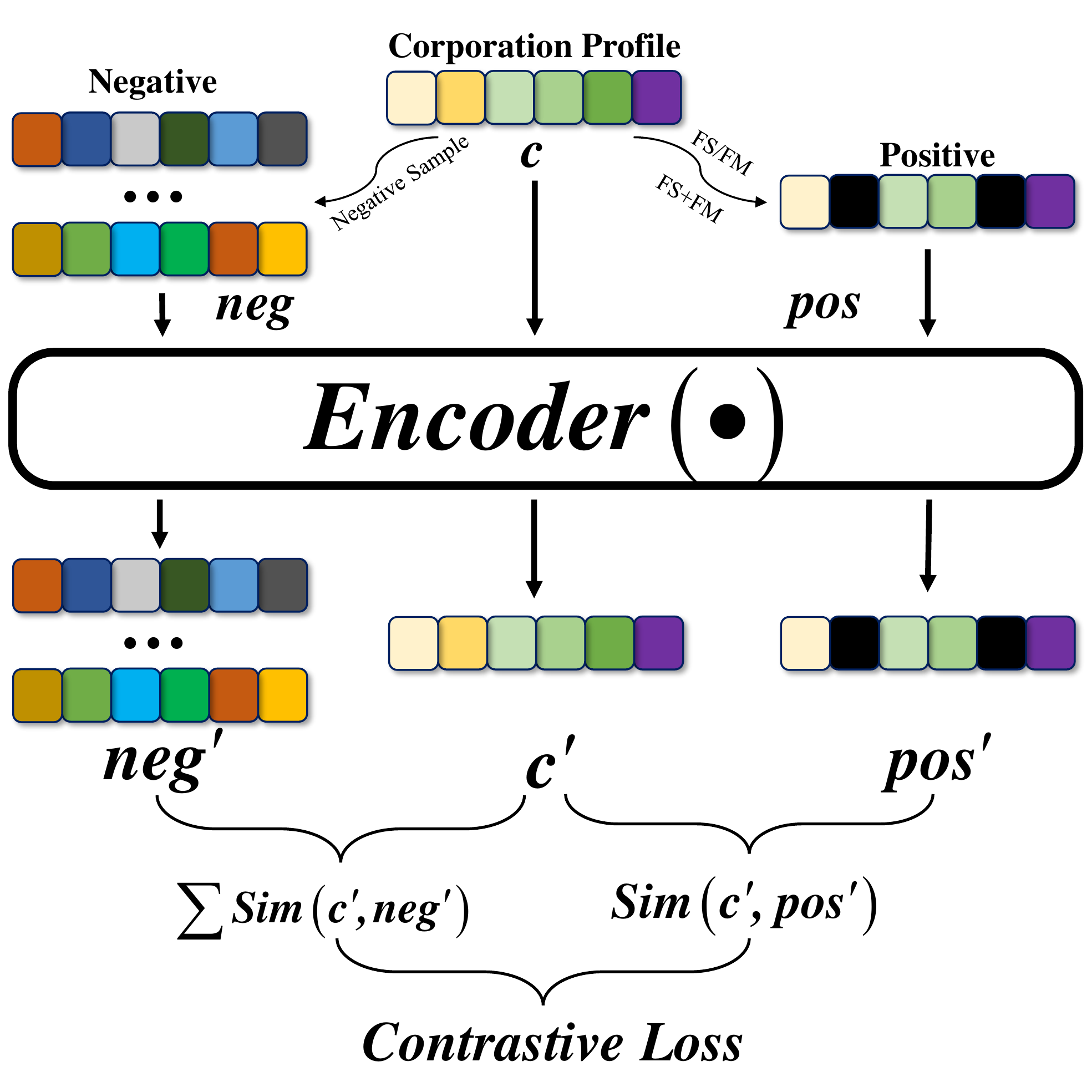}  
\caption{The architecture of the Contrastive Self-superivised Pre-training.} 
\label{figure2}
\end{figure}

The architecture of the Contrastive Self-supervised Pre-training is described as Figure \ref{figure2}. Positive samples and negative samples are generated by self-supervised learning tasks and negative sampling, respectively. They are all mapped into another vector space through an encoder function $Encoder(\cdot)$ to get better initialization. Finally, the encoder model is trained by constructing the contrastive loss between positive samples and negative samples.

\noindent
\textit{\textbf{Self-supervised Learning Tasks.}} Inspired by masked-LM used in pre-training for NLP and unordered nature of corporation profile, we propose two methods for generating positive samples: Feature Masking (FM), Feature Swapping (FS), and design three self-supervised learning tasks: FS, FM, FS+FM. During the FS task, we can randomly swap the location of features for $@k$ times due to the unordered nature of the corporation profile to build the self-supervised task. In addition, as demonstrated in figure \ref{figure2}, FM is designed to mask a subset of input features, therefore, they can only train by partial features $@k$. FS+FM is a more complicated task by using FS and FM simultaneously.
\\

\noindent
\textit{\textbf{Encoder Function.}} Theoretically, any encoder function which could map a vector space into another vector space can be applied to our framework. In our experiments, we use MLP and Transformer. MLP is easy and simple while Transformer is more powerful but more complicated. It is defined as follows:
\begin{equation}
    c^{\prime} = Encoder(c)
\end{equation}
where $c^{\prime}\in \mathbb{R}^{d^{\prime}}$, $c^{\prime}$ is the representation of corporation in new vector space, $d^{\prime}$ is the dimension of new vector space.
\\

\noindent
\textit{\textbf{Negative Samples and Contrastive Loss.}} For negative samples, we can randomly select $@L$ sample which different from corresponding positive samples. Here we use cosine similarity to measure the distance between the embeddings of two samples. And InfoNce loss is applied to build contrastive loss. The \textit{self-supervised pre-training }(SSP) loss can be represented as follows:
\begin{equation}
    sim(u,v)=\frac{u\cdot v}{ \Vert u\Vert \Vert v\Vert}
\end{equation}
\begin{equation}
    \mathcal{L}_{ssp} = -log\frac{\exp \left(\frac{sim(c^{\prime},pos^{\prime})}{\tau}\right)}{\exp \left(\frac{sim(c^{\prime},pos^{\prime})}{\tau}\right) + \sum_{l=0}^L\exp \left(\frac{sim(c^{\prime},neg_l^{\prime})}{\tau}\right)}
\end{equation}
\subsection{Vector Space Stacking}
To fully use the information of data, we propose two methods of vector space stacking: Space Concatenation (SC) and Space Fusion (SF). They can be formulated as follows:
\begin{equation}
    x_{sc} = Concatenation(c,c^{\prime})
\end{equation}
\begin{equation}
    x_{sf} = \alpha c +(1-\alpha)c^{\prime}
\end{equation}
where $x_{sc}\in \mathbb{R}^{d+d^{\prime}}$, and $x_{fc}\in \mathbb{R}^{d}$ due to $d=d^{\prime}$.

When the new vector space and origin vector space are in different space, in other words, $d^{\prime}\neq d$, we can concatenate them together to form a large space. In this space, both origin data and pre-trained data can be used to predict the credit level. What's more, if $d^{\prime} = d$, not only Space Concatenation but also Space Fusion can be used to mix two spaces. When $\alpha=1$ CP4CCR will degenerate into the common credit rating model. While $\alpha=1$, only pre-trained data will be used. The reason why we design this structure will be explained in Section \ref{section:Experiment} (Experiment).
\subsection{Phase 2}
After the pre-training in phase 1, the better initialization of data will be got. In this vector space, similar samples will be closer, and dissimilar samples will be far from each other. The border between different classes will be wider. Therefore, this distribution of data will be easy for standard credit rating models to learn, especially for the class with few samples. The result of experiments will demonstrate this point. The training process can be formulated as follows.
\begin{equation}
    z =CreditRatingModel(x) 
\end{equation}
\begin{equation}
    \hat{y} = log\_softmax(z)
\end{equation}
\begin{equation}
    \mathcal{L}(\hat{y})=-\sum\limits_{i=1}^{n}{{{y}_{i}}\log ({{{\hat{y}}}_{i}})+(1-{{y}_{i}})\log (1-{{{\hat{y}}}_{i}})}+\lambda {{\left\| \Delta  \right\|}^{2}}
\end{equation}
where $y$ denotes the one-hot encoding vector of ground truth item, $\lambda$ is parameter-specific regularization hyperparameters to prevent overfitting, and the model parameters are $\Delta$. The Back-Propagation algorithm is performed to train the model.
\section{Experiment}\label{section:Experiment}
In this section, we aim to answer the following three questions:

\textbf{RQ1.} Could the proposed CP4CCR improve the performance of standard credit rating models?

\textbf{RQ2.} How the proposed CP4CCR infect the data distribution of class with few samples?

\textbf{RQ3.} How do different types of vector space stacking affect the model performance?
\subsection{Experimental Configurations}
\textit{\textbf{Dataset.}} We evaluate the proposed CP4CCR on the corporate credit dataset. It has been built based on the annual financial statements of Chinese listed companies and China Stock Market \& Accounting Research Database (CSMRA). The results of credit ratings are conducted by the famous credit rating agencies, including CCXI, China Lianhe Credit Rating (CLCR), etc. The financial data of the corporation includes six aspects: profit capability, operation capability, growth capability, repayment capability, cash flow capability, Dupont identity. After the same preprocess as work \cite{feng2020corporation},we get 39 features and 9 rating labels: AAA, AA, A, BBB, BB, B, CCC, CC, C.
\\

\noindent
\textit{\textbf{Baselines.}} To evaluate the performance of the proposed CP4CCR, we select several models including KNN, Logistic Regression (LR), Random Forest (RF), Decision Tree (DT), GBDT, AdaBoost, GaussianNB, LDA, SVM (linear), SVM (rbf), MLP and Xgboost due to the particularity of the financial field. But we believe our framework will have more or less improvement based on other corporate credit rating models.
\\

\noindent
\textit{\textbf{Evaluation Metrics.}} We adopt three commonly-used metrics for evaluation, including Recall, Accuracy, and F1-score. Let $\mathcal{R}$ denote the set of credit rating models, and $\mathcal{M}$ is the set of metrics. For comparing overall improvement on all kinds of rating models, we propose a metric, \underline{O}ver\underline{a}ll \underline{P}erformance (OaP), which can be computed as follows:
\begin{equation}
    OaP = \sum_{r \in \{\mathcal{R}\}} \sum_{t \in \{\mathcal{M}\}} {SSP(r,t)-Origin(r,t)}
\end{equation}
where $SSP(r,t)$ means the metric $t$ of model $r$ under self-supervised pre-training, while $Origin(r,t)$ uses the origin dataset. On the whole, $OaP>0$ means CP4CCR has a positive gain on all models, negative gain vice versa.
\\

\noindent
\textit{\textbf{Hyperparameter Setup.}} During the data preprocessing, we perform the Feature Masking, Feature Swapping and FS+FM respectively. $k$ is set to 4, and negative samples $L$ is set to 8. In the experiment, we use the MLP or Transformer as the encoder function.  The initial learning rate for Adam is set to 0.001 and will decay by 0.00001.

\subsection{Improvement with Baseline Methods (RQ1)}

\begin{table}[!hbtp]
\centering
\caption{The Overall Performance of Experiments.}
\label{table2}
\begin{tabular}{ccc}
\toprule[1.5pt]
                    & { \textbf{MLP}} & {\textbf{Transformer}} \\ \hline
 { \textbf{FS}}     & 0.60619 & 1.01328\\
 { \textbf{FM}}     & 0.81989 & 1.03799\\
 { \textbf{FS+FM}}  & 0.86575 & 1.04080\\
\toprule[1.5pt]
\end{tabular}

\end{table}

To demonstrate the improvement of our proposed framework CP4CCR, we conduct experiments based on the baseline methods. The origin denotes the data without pre-training. Data are pre-trained by three different self-supervised tasks:@FS, @FM, @FS+FM. We find that most baselines are improved by pre-training. Due to the space limit, we omit the sophisticated original result.  For clearly verifying, the overall performance (OaP) is displayed in Table \ref{table2}. Both two encoder functions have a positive gain for data pre-training while the transformer encoder is better. FM is superior to FS, and perform them simultaneously will bring better results.

Besides we find another interesting result, in terms of the powerful encoder (Transformer), different types of pretext tasks bring almost the same positive gain for the model. However, a normal encoder (MLP) has a relative gap between different pretext tasks. It is might that a more powerful encoder is more robust to the pretext tasks. Therefore, we should pay more attention to the design of pretext tasks for the normal encoder, while for the powerful encoder, concentrating on improving the ability of it is a better way.  

\subsection{Influence on the Class Imbalance(RQ2)}
To clearly demonstrate the influence on the class-imbalanced problem, we perform the t-SNE visualization on the original and pre-trained data. 

\begin{figure}[htbp]  
\centering  
\includegraphics[scale=0.25]{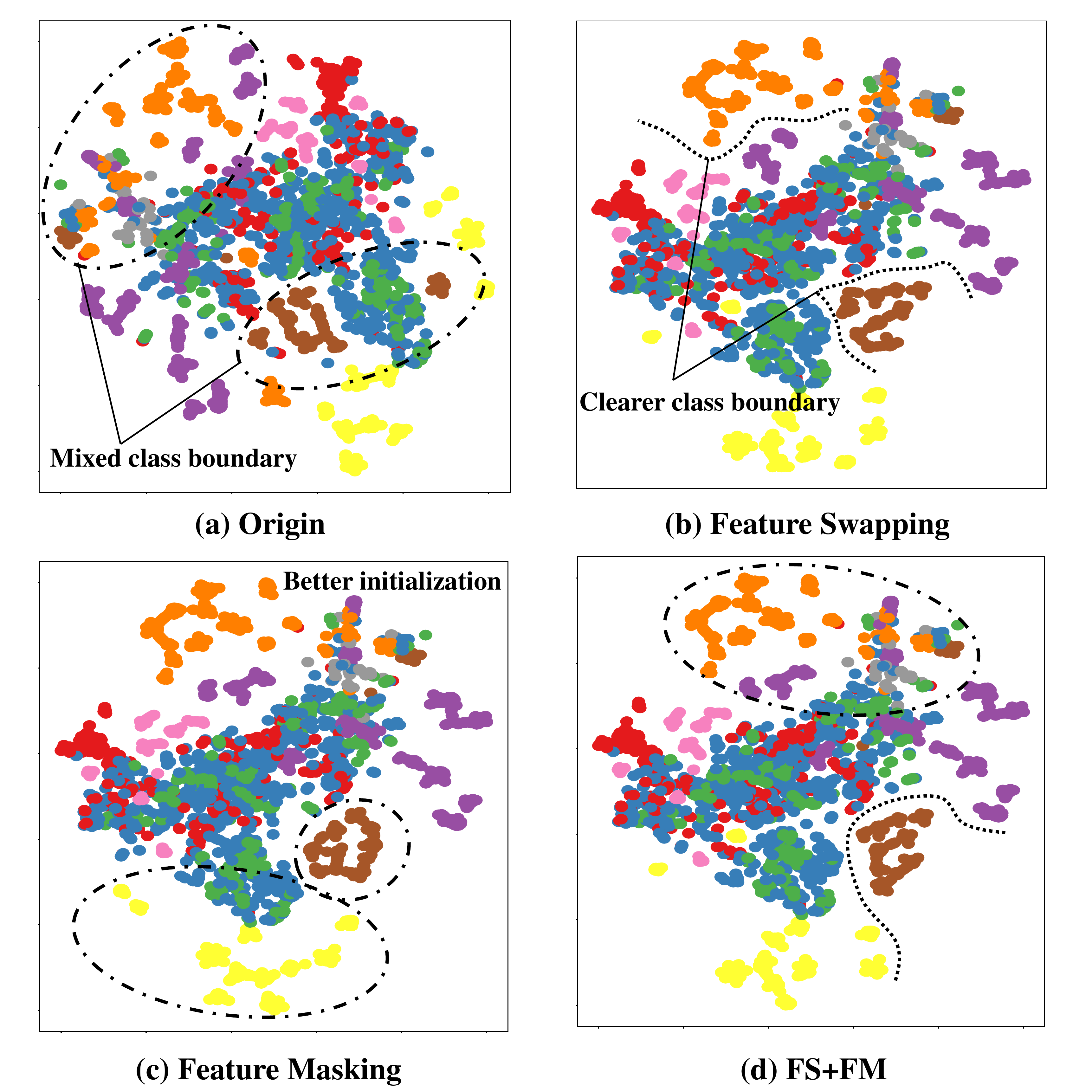}  
\caption{t-SNE visualization of origin data and pretrained data by FS, FM and FS+FM} 
\label{figure3}
\end{figure}
From Figure \ref{figure3}, the original data is usually chaotic. But after contrastive self-supervised learning, the imbalanced problem is mitigated, which results in clearer boundary and better initialization. For example, we can find that there are clear boundaries with feature swapping self-supervised tasks from Figure \ref{figure3} (b). And in Figure \ref{figure3} (c) yellow and brown samples get better initialization compared with original data to easily perform downstream tasks (Corporate Credit Rating).
\subsection{Different Vector Space Stackings (RQ3)}
To answer the \textbf{RQ3}, in this section, we analyze how hyper-parameter $\alpha$ impacts the metric OvP when the dimensions of origin and new vector space are the same, in other words, $d=d^{\prime}$. The following Figure \ref{figure4} shows the result. 

\begin{figure}[htbp]  
\centering  
\includegraphics[scale=0.28]{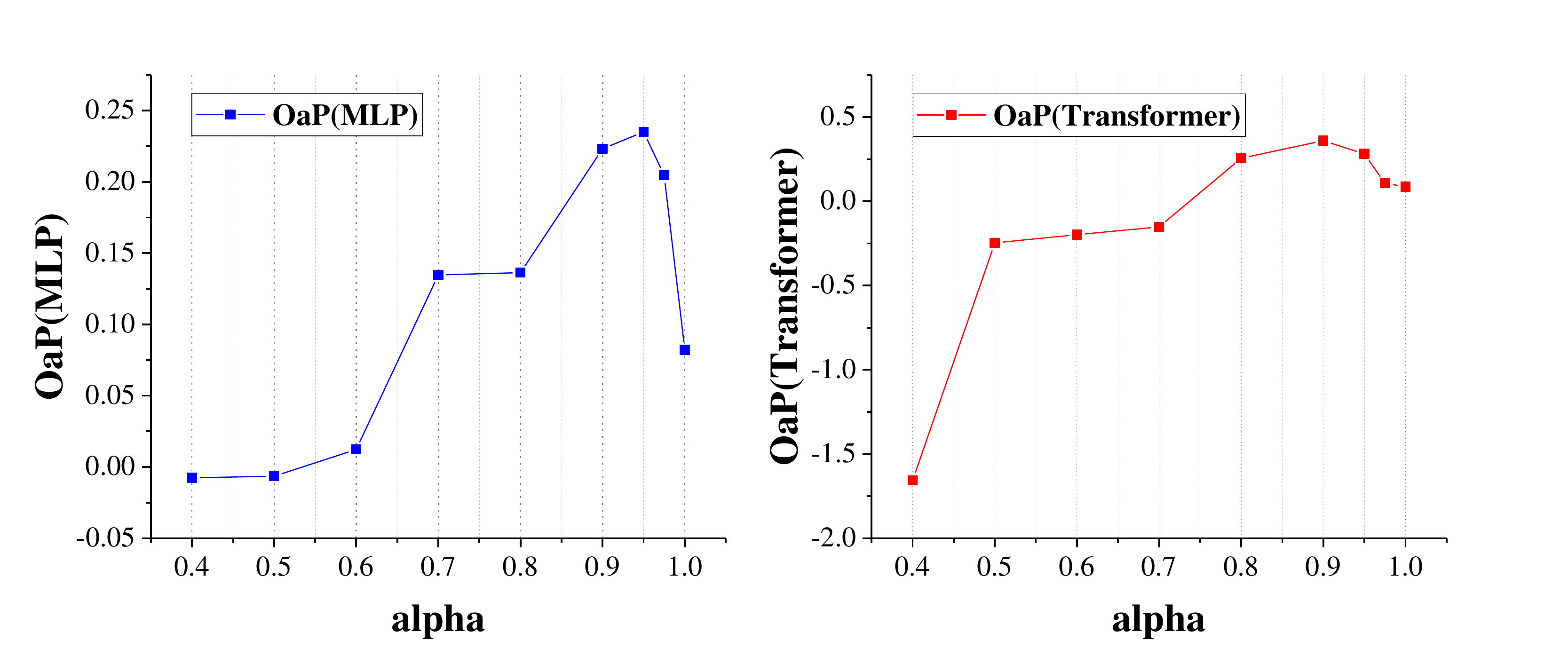}  
\caption{The influence on the Space Fusion} 
\label{figure4}
\end{figure}
We can find that when $\alpha$ is too small, space fusion even results in a negative gain. It gets better performance when $\alpha$ is about 0.9. And in the whole, the MLP encoder is stable, while the transformer encoder performs better but bigger variance.

\begin{figure}[htbp]  
\centering  
\includegraphics[scale=0.2]{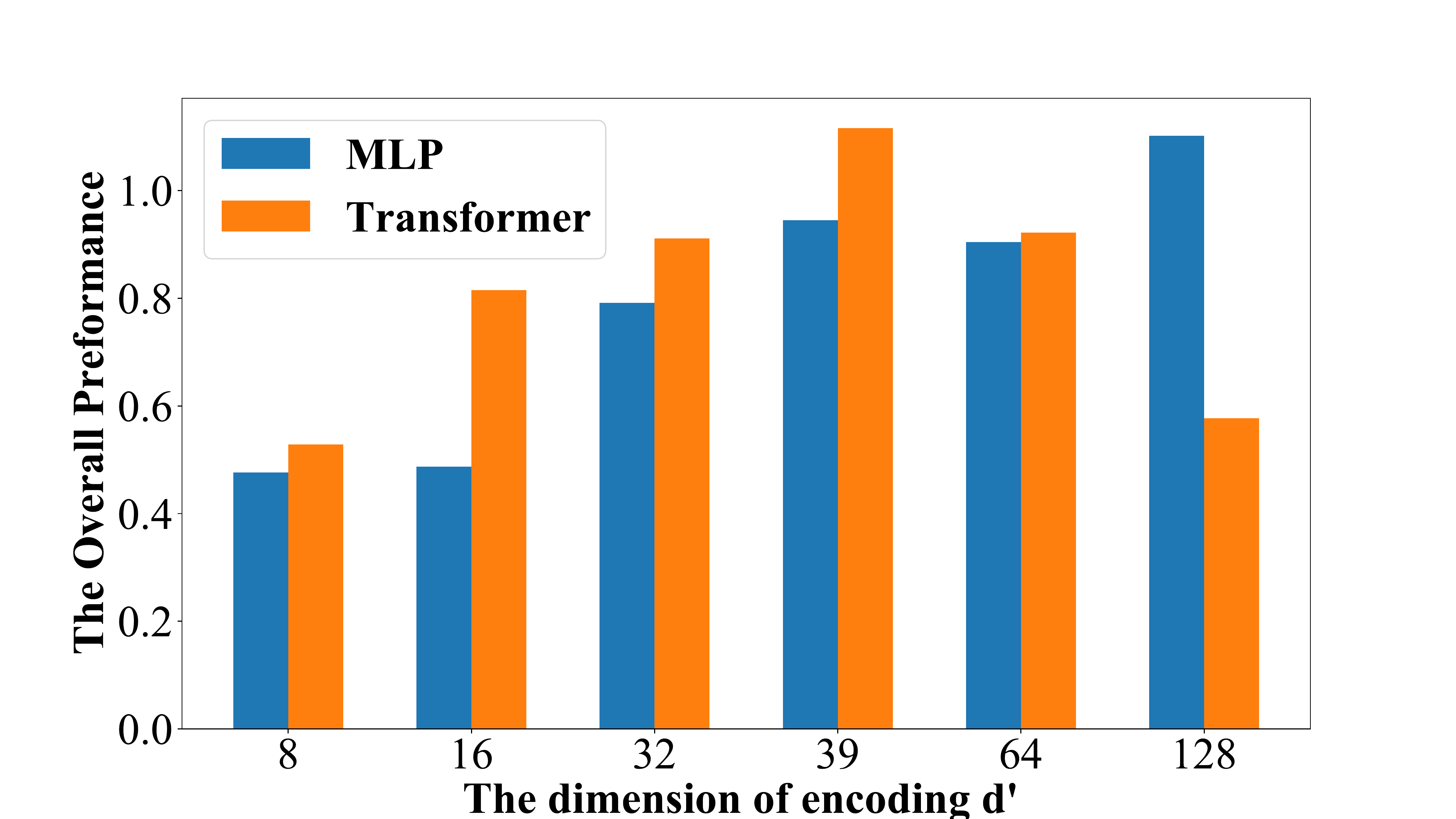}  
\caption{The influence on the Space Concatenation} 
\label{figure5}
\end{figure}

What's more, we analyze the influence on space concatenation when $d\neq d^{\prime}$, the variation of OaP follows the changes of encoding dimension $d^{\prime}$. From Figure \ref{figure5}, we can draw some conclusions. When $d^{\prime}$ is less than 64, the Transformer encoder performs better than MLP. As $d^{\prime}$ increasing, larger vector space brings more positive gain in MLP than Transformer. We believe that increasing $d^{\prime}$ leads to more parameters for Transformer encoder which converges difficultly. In addition to considering the time of cost, we believe that a relatively satisfied result is got when $d=d^{\prime}$ with space concatenation.

All in all, both FS and FM have positive gain to get better initialization and clearer boundary. Then Transformer is better than MLP to be the encoder function in the majority of cases. Finally, in terms of different space stackings, space concatenation exactly a great way to mix the original and new data information.

\section{Conclusion}\label{section:Conclusion}
In this work, we propose a novel framework named Contrastive Pre-training for Imbalanced Corporate Credit Rating (CP4CCR), which can get better initialization and clearer boundary by contrastive self-supervised learning. In specific, we proposed two self-supervised tasks: Feature Masking and Feature Swapping, and two types of vector space stacking: Space Concatenation and Space Fusion. Extensive experiments demonstrate that our CP4CCR can improve the performance of baseline methods by pre-trained data. What's more, Transformer is a better encoder function for contrastive self-supervised pre-training. In terms of space stackings, space concatenation is still a better way. We believe this work will bring a new framework to tackle imbalanced corporate credit rating in reality.

\bibliographystyle{ACM-Reference-Format}
\bibliography{sample-base}

\begin{thebibliography}{}

\bibitem[\protect\citeauthoryear{Bruss \bgroup \em et al.\egroup
  }{2019}]{bruss2019deeptrax}
C~Bayan Bruss, Anish Khazane, Jonathan Rider, Richard Serpe, Antonia Gogoglou,
  and Keegan~E Hines.
\newblock Deeptrax: Embedding graphs of financial transactions.
\newblock {\em arXiv preprint arXiv:1907.07225}, 2019.

\bibitem[\protect\citeauthoryear{Caron \bgroup \em et al.\egroup
  }{2020}]{caron2020unsupervised}
Mathilde Caron, Ishan Misra, Julien Mairal, Priya Goyal, Piotr Bojanowski, and
  Armand Joulin.
\newblock Unsupervised learning of visual features by contrasting cluster
  assignments.
\newblock {\em arXiv preprint arXiv:2006.09882}, 2020.

\bibitem[\protect\citeauthoryear{Chen and Long}{2020}]{chen2020novel}
Binbin Chen and Shengjie Long.
\newblock A novel end-to-end corporate credit rating model based on
  self-attention mechanism.
\newblock {\em IEEE Access}, 8:203876--203889, 2020.

\bibitem[\protect\citeauthoryear{Chen \bgroup \em et al.\egroup
  }{2020}]{chen2020simple}
Ting Chen, Simon Kornblith, Mohammad Norouzi, and Geoffrey Hinton.
\newblock A simple framework for contrastive learning of visual
  representations.
\newblock {\em arXiv preprint arXiv:2002.05709}, 2020.

\bibitem[\protect\citeauthoryear{Cheng \bgroup \em et al.\egroup
  }{2019a}]{cheng2019risk}
Dawei Cheng, Yi~Tu, Zhen-Wei Ma, Zhibin Niu, and Liqing Zhang.
\newblock Risk assessment for networked-guarantee loans using high-order graph
  attention representation.
\newblock In {\em IJCAI}, pages 5822--5828, 2019.

\bibitem[\protect\citeauthoryear{Cheng \bgroup \em et al.\egroup
  }{2019b}]{cheng2019dynamic}
Dawei Cheng, Yiyi Zhang, Fangzhou Yang, Yi~Tu, Zhibin Niu, and Liqing Zhang.
\newblock A dynamic default prediction framework for networked-guarantee loans.
\newblock In {\em Proceedings of the 28th ACM International Conference on
  Information and Knowledge Management}, pages 2547--2555, 2019.

\bibitem[\protect\citeauthoryear{Cheng \bgroup \em et al.\egroup
  }{2020a}]{cheng2020contagious}
Dawei Cheng, Zhibin Niu, and Yiyi Zhang.
\newblock Contagious chain risk rating for networked-guarantee loans.
\newblock In {\em Proceedings of the 26th ACM SIGKDD International Conference
  on Knowledge Discovery \& Data Mining}, pages 2715--2723, 2020.

\bibitem[\protect\citeauthoryear{Cheng \bgroup \em et al.\egroup
  }{2020b}]{cheng2020spatio}
Dawei Cheng, Sheng Xiang, Chencheng Shang, Yiyi Zhang, Fangzhou Yang, and
  Liqing Zhang.
\newblock Spatio-temporal attention-based neural network for credit card fraud
  detection.
\newblock In {\em Proceedings of the AAAI Conference on Artificial
  Intelligence}, volume~34, pages 362--369, 2020.

\bibitem[\protect\citeauthoryear{Chi \bgroup \em et al.\egroup
  }{2020}]{chi2020infoxlm}
Zewen Chi, Li~Dong, Furu Wei, Nan Yang, Saksham Singhal, Wenhui Wang, Xia Song,
  Xian-Ling Mao, Heyan Huang, and Ming Zhou.
\newblock Infoxlm: An information-theoretic framework for cross-lingual
  language model pre-training.
\newblock {\em arXiv preprint arXiv:2007.07834}, 2020.

\bibitem[\protect\citeauthoryear{Fang and Xie}{2020}]{fang2020cert}
Hongchao Fang and Pengtao Xie.
\newblock Cert: Contrastive self-supervised learning for language
  understanding.
\newblock {\em arXiv preprint arXiv:2005.12766}, 2020.

\bibitem[\protect\citeauthoryear{Feng \bgroup \em et al.\egroup
  }{2020a}]{feng2020graph}
Bojing Feng, Haonan Xu, Wenfang Xue, and Bindang Xue.
\newblock Every corporation owns its structure: Corporate credit ratings via
  graph neural networks.
\newblock {\em arXiv preprint arXiv:2012.01933}, 2020.

\bibitem[\protect\citeauthoryear{Feng \bgroup \em et al.\egroup
  }{2020b}]{feng2020corporation}
Bojing Feng, Haonan Xu, Wenfang Xue, and Bindang Xue.
\newblock Every corporation owns its structure: Corporate credit ratings via
  graph neural networks, 2020.

\bibitem[\protect\citeauthoryear{Feng \bgroup \em et al.\egroup
  }{2020c}]{feng2020every}
Bojing Feng, Wenfang Xue, Bindang Xue, and Zeyu Liu.
\newblock Every corporation owns its image: Corporate credit ratings via
  convolutional neural networks.
\newblock {\em arXiv preprint arXiv:2012.03744}, 2020.

\bibitem[\protect\citeauthoryear{Giorgi \bgroup \em et al.\egroup
  }{2020}]{giorgi2020declutr}
John~M Giorgi, Osvald Nitski, Gary~D Bader, and Bo~Wang.
\newblock Declutr: Deep contrastive learning for unsupervised textual
  representations.
\newblock {\em arXiv preprint arXiv:2006.03659}, 2020.

\bibitem[\protect\citeauthoryear{Gogas \bgroup \em et al.\egroup
  }{2014}]{gogas2014forecasting}
Periklis Gogas, Theophilos Papadimitriou, and Anna Agrapetidou.
\newblock Forecasting bank credit ratings.
\newblock {\em The Journal of Risk Finance}, 2014.

\bibitem[\protect\citeauthoryear{Golbayani \bgroup \em et al.\egroup
  }{2020}]{golbayani2020application}
Parisa Golbayani, Dan Wang, and Ionut Florescu.
\newblock Application of deep neural networks to assess corporate credit
  rating.
\newblock {\em arXiv preprint arXiv:2003.02334}, 2020.

\bibitem[\protect\citeauthoryear{He \bgroup \em et al.\egroup
  }{2020}]{he2020momentum}
Kaiming He, Haoqi Fan, Yuxin Wu, Saining Xie, and Ross Girshick.
\newblock Momentum contrast for unsupervised visual representation learning.
\newblock In {\em Proceedings of the IEEE/CVF Conference on Computer Vision and
  Pattern Recognition}, pages 9729--9738, 2020.

\bibitem[\protect\citeauthoryear{Jaiswal \bgroup \em et al.\egroup
  }{2020}]{jaiswal2020survey}
Ashish Jaiswal, Ashwin~Ramesh Babu, Mohammad~Zaki Zadeh, Debapriya Banerjee,
  and Fillia Makedon.
\newblock A survey on contrastive self-supervised learning.
\newblock {\em arXiv preprint arXiv:2011.00362}, 2020.

\bibitem[\protect\citeauthoryear{Jaiswal \bgroup \em et al.\egroup
  }{2021}]{jaiswal2021survey}
Ashish Jaiswal, Ashwin~Ramesh Babu, Mohammad~Zaki Zadeh, Debapriya Banerjee,
  and Fillia Makedon.
\newblock A survey on contrastive self-supervised learning.
\newblock {\em Technologies}, 9(1):2, 2021.

\bibitem[\protect\citeauthoryear{Kim}{2005}]{kim2005predicting}
Kee~S Kim.
\newblock Predicting bond ratings using publicly available information.
\newblock {\em Expert Systems with Applications}, 29(1):75--81, 2005.

\bibitem[\protect\citeauthoryear{Mikolov \bgroup \em et al.\egroup
  }{2013}]{mikolov2013distributed}
Tomas Mikolov, Ilya Sutskever, Kai Chen, Greg~S Corrado, and Jeff Dean.
\newblock Distributed representations of words and phrases and their
  compositionality.
\newblock In {\em Advances in neural information processing systems}, pages
  3111--3119, 2013.

\bibitem[\protect\citeauthoryear{Misra and Maaten}{2020}]{misra2020self}
Ishan Misra and Laurens van~der Maaten.
\newblock Self-supervised learning of pretext-invariant representations.
\newblock In {\em Proceedings of the IEEE/CVF Conference on Computer Vision and
  Pattern Recognition}, pages 6707--6717, 2020.

\bibitem[\protect\citeauthoryear{Pai \bgroup \em et al.\egroup
  }{2015}]{pai2015credit}
Ping-Feng Pai, Yi-Shien Tan, and Ming-Fu Hsu.
\newblock Credit rating analysis by the decision-tree support vector machine
  with ensemble strategies.
\newblock {\em International Journal of Fuzzy Systems}, 17(4):521--530, 2015.

\bibitem[\protect\citeauthoryear{Petropoulos \bgroup \em et al.\egroup
  }{2016}]{petropoulos2016novel}
Anastasios Petropoulos, Sotirios~P Chatzis, and Stylianos Xanthopoulos.
\newblock A novel corporate credit rating system based on student’st hidden
  markov models.
\newblock {\em Expert Systems with Applications}, 53:87--105, 2016.

\bibitem[\protect\citeauthoryear{Trinh \bgroup \em et al.\egroup
  }{2019}]{trinh2019selfie}
Trieu~H Trinh, Minh-Thang Luong, and Quoc~V Le.
\newblock Selfie: Self-supervised pretraining for image embedding.
\newblock {\em arXiv preprint arXiv:1906.02940}, 2019.

\bibitem[\protect\citeauthoryear{Wu and Hsu}{2012}]{wu2012credit}
Tsui-Chih Wu and Ming-Fu Hsu.
\newblock Credit risk assessment and decision making by a fusion approach.
\newblock {\em Knowledge-Based Systems}, 35:102--110, 2012.

\bibitem[\protect\citeauthoryear{Xie \bgroup \em et al.\egroup
  }{2020}]{xie2020contrastive}
Xu~Xie, Fei Sun, Zhaoyang Liu, Jinyang Gao, Bolin Ding, and Bin Cui.
\newblock Contrastive pre-training for sequential recommendation.
\newblock {\em arXiv preprint arXiv:2010.14395}, 2020.

\bibitem[\protect\citeauthoryear{Yang and Xu}{2020}]{yang2020rethinking}
Yuzhe Yang and Zhi Xu.
\newblock Rethinking the value of labels for improving class-imbalanced
  learning.
\newblock {\em Advances in Neural Information Processing Systems}, 33, 2020.

\bibitem[\protect\citeauthoryear{Yao \bgroup \em et al.\egroup
  }{2020}]{Yao2020SelfsupervisedLF}
Tiansheng Yao, Xinyang Yi, D.~Cheng, F.~Yu, Ting Chen, Aditya Menon, L.~Hong,
  Ed~Huai hsin Chi, Steve Tjoa, J.~Kang, and Evan Ettinger.
\newblock Self-supervised learning for large-scale item recommendations.
\newblock {\em arXiv: Learning}, 2020.

\bibitem[\protect\citeauthoryear{Yeh \bgroup \em et al.\egroup
  }{2012}]{yeh2012hybrid}
Ching-Chiang Yeh, Fengyi Lin, and Chih-Yu Hsu.
\newblock A hybrid kmv model, random forests and rough set theory approach for
  credit rating.
\newblock {\em Knowledge-Based Systems}, 33:166--172, 2012.

\end{thebibliography}

\end{document}